\documentclass{article} 
\usepackage{aditi}

\usepackage{microtype}
\usepackage{hyperref}
\usepackage{url}
\usepackage{booktabs}
\usepackage{amsfonts}       
\usepackage{nicefrac}       
\usepackage{xcolor}         
\usepackage{bm}             
\usepackage{graphicx}       
\usepackage{amsmath}        
\usepackage{amssymb}        
\usepackage{amsthm}         
\usepackage{xspace}         
\usepackage{adjustbox}      
\usepackage{multirow}       
\usepackage{pifont}         
\usepackage{caption}        
\usepackage{listings}       
\usepackage{fancyvrb}       
\usepackage{inconsolata}                                    
\usepackage{wrapfig}        
\usepackage{enumitem}

\usepackage{style} 
\definecolor{skyblue}{RGB}{204,229,255}

\usepackage[most]{tcolorbox}
\tcbset{
  lightbluebox/.style={
    colback=skyblue!70,        
    colframe=skyblue!75!black, 
    boxrule=1pt,
    arc=4pt,
    auto outer arc
  }
}

\definecolor{darkblue}{rgb}{0, 0, 0.5}
\hypersetup{colorlinks=true, citecolor=darkblue, linkcolor=darkblue, urlcolor=darkblue}

\setTitleruleGap{0.75pt}         

\title{Mitigating Modal Imbalance in Multimodal Reasoning}

\setauthors{Chen Henry Wu$^{\star, 1}$ \authorsep Neil Kale$^{\star, 1}$ \authorsep Aditi Raghunathan$^1$}
\setaffils{$^{1}$ Carnegie Mellon University \affilsep}

\setemail{\{chenwu2,nkale,aditirag\}@cs.cmu.edu}
\setcode{https://github.com/AR-FORUM/modal-imbalance}
\setwebsite{https://ar-forum.github.io/modalimbalanceweb/}

\usepackage{eqparbox}                                       
\newcommand{\gptfo}{\text{GPT-4o}\xspace}                   
\newcommand{\dalle}{\text{DALL-E}\xspace}                   
\newcommand{\lm}{\text{FM}\xspace}                          

\begin{document}

\maketitle

\begin{abstract}
    Foundation models (\lm{}s) deployed in real-world tasks such as computer-use agents must integrate diverse modalities. How good are \lm{}s at performing \textit{joint reasoning}, simultaneously reasoning over multiple modalities, especially when the modalities interact and relate to each other to form \textit{cross-modal context}? To better understand this problem, we study \lm{}s on \textit{cross-modal conflicts}: scenarios where conflicting evidence is presented across modalities. This allows us to examine whether \lm{}s prioritize one modality over another or reason jointly to reconcile the conflict.
    Our experiments reveal that \lm{}s can recognize conflicts in \textit{unimodal contexts}, composed of a single modality, 90\% of the time, but the ratio falls as low as 3\% when evidence is split across modalities -- similar observations hold in \textit{cross-lingual contexts}, composed of multiple languages. We trace this failure to \textit{cross-modal attention imbalance}, showing that \lm{}s exhibit extreme asymmetry in attention scores, disproportionately prioritizing certain modalities. We show that cross-modal attention imbalance does not go away by simply scaling up multimodal or multilingual datasets blindly, since they lack training examples that explicitly require cross-modal reasoning. We demonstrate that even a simple and scalable method of explicitly combining multiple modalities within each training instance significantly reduces attention imbalance. Reduced attention imbalance directly translates to improved downstream performance on several vision-language benchmarks. Our findings underscore the importance of systematically addressing cross-modal contexts to build reliable foundation models.
\end{abstract}

\section{Introduction}

Recent advances in foundation models (\lm{}s; \citealp{gpt4o,geminiteam2024gemini,claude3}) have enabled their deployment in increasingly complex tasks that require reasoning over diverse information sources. From autonomous web browsing \citep{act-1,anthropic2024computeruse, openai2024operator} to AI-driven research assistants \citep{perplexity2024prosearch, sakana2024aiscientist}, \lm{}s are now tasked with reasoning jointly over multiple domains such as text, images, code, and structured data.

\begin{figure*}[!th]
\centering
    \includegraphics[width=1.0\linewidth]{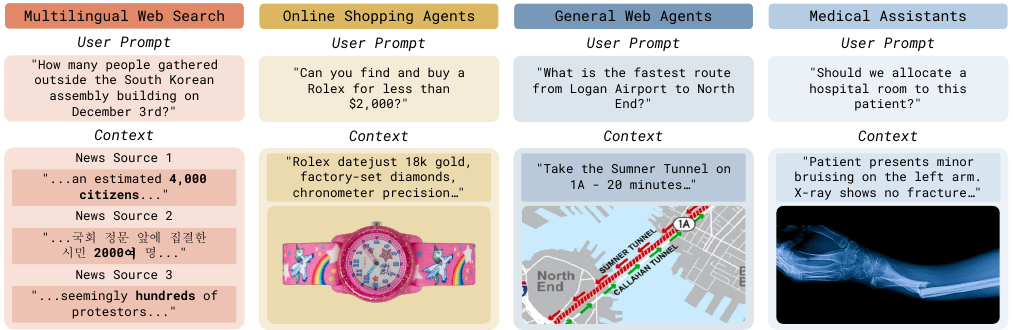}
    \caption{\label{fig:motivation} 
    \lm{}-based agents need to reason over diverse modalities, such as multilingual news, online shopping websites, maps, and EHR records. Failure to handle cross-modal context can result in consequences including misinformation (orange), purchasing a scam (yellow), misdirection (blue), or even providing the wrong medical treatment (light blue).}
    \vspace{-10pt}
\end{figure*}

However, existing work indicates that \lm{}s fall short when handling inputs from non-textual modalities. For example, some studies demonstrate that \lm{}s answer visual questions primarily based on language priors, disregarding actual visual inputs \citep{Winterbottom2020OnMB,Niu2020CounterfactualVA,Lin2023RevisitingTR}; others illustrate scenarios where models hallucinate objects absent from the image during open-ended generation tasks \citep{Sun2023AligningLM}. Yet, it remains unclear whether this behavior originates primarily from a context-parametric gap \citep{goyal2024context} or a modality gap \citep{Liang2022MindTG}.

In this work, we specifically focus on the capability of \lm{}s to reason across modalities when all necessary information is explicitly provided in the input context and does not conflict with parametric knowledge. This setup is especially relevant for \lm{} agents and assistants that need to interpret up-to-date information unavailable within their parametric knowledge -- such as web pages combining images, multilingual text, and embedded scripts (Figure \ref{fig:motivation}). By designing scenarios that isolate cross-modal reasoning from parametric knowledge retrieval, we directly assess how effectively these models reason over multiple modalities. 

As a clean and concrete test case for this, we create \textit{cross-modal conflict} datasets where each modality provides contrasting evidence. This allows us to examine whether \lm{}s prioritize one modality over another or reason jointly to reconcile the conflict. Joint reasoning means relating multiple pieces of information and identifying insights that can only be obtained from their interaction, not from any single piece alone. In our setup, joint reasoning over the contrasting evidence ideally identifies that there is no clear answer since the evidence is in direct conflict. Our experiments reveal a striking gap: while \lm{}s perform well in \textit{unimodal} settings (e.g., text-text or image-image), their ability to detect conflicts deteriorates significantly by up to 65\% in \textit{cross-modal} contexts (e.g., text-image). 
Moreover, this degradation extends to multilingual scenarios, where monolingual performance (e.g., English-English or Chinese-Chinese) is significantly better than multilingual performance (e.g., English-Chinese). 

We investigate what drives this behavior, where state-of-the-art models exclusively rely on evidence from one modality rather than jointly reasoning. First, we observe that it is not simply a consequence of models being weak in one modality (\S\ref{subsec:result}). VLMs detect conflicts between multiple images as easily as conflicts between multiple texts (Figure \ref{fig:heterogeneity}). This extends to multilingual settings -- conflicts between multiple Chinese texts are detected as often as in English.

We hypothesize that the gap between unimodal and cross-modal conflict detection is because of \textit{cross-modal attention imbalance}: an extreme asymmetry in attention scores, where \lm{}s disproportionately prioritize certain
modalities. We validate our hypothesis by finding that manual attention reweighting vastly shifts the model towards joint reasoning rather than relying on one modality over another. Manual attention reweighting also directly improves downstream performance on a variety of hard vision-language benchmarks (\S\ref{sec:attention-imbalance}).

\begin{figure*}[!t]
\centering
    \includegraphics[width=\linewidth]{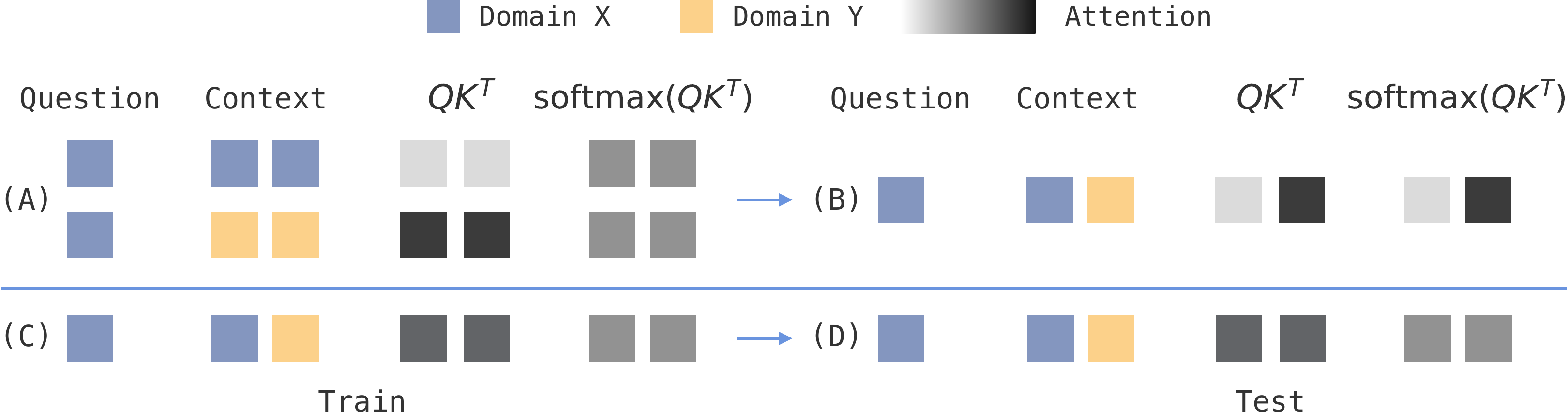}
    \caption{\label{fig:illustration} \textbf{An illustration for cross-modal attention imbalance.} In unimodal contexts (A), different domains show balanced normalized attention ($\mathrm{softmax}(QK^\top)$) despite divergent pre-softmax logits ($QK^\top$). Cross-modal contexts (B) expose cross-modal attention imbalance -- normalization fails to mitigate logit-level imbalance. Instance-level modality mixing (C) resolves this by training models to intrinsically balance attention logits across modalities.}
\end{figure*}

We investigate how to correct cross-modal attention imbalance. The problem is not resolved by simply adding more training data in each modality (\S\ref{subsec:dataset-level-mixing-fails}). As illustrated in Figure~\ref{fig:illustration}, 
when the cross-modal attention scores are imbalanced, different modalities have different pre-softmax logits ($QK^\top$). However, after normalization, unimodal contexts show balanced normalized attention ($\mathrm{softmax}(QK^\top)$). So, fine-tuning on either modality separately does not reduce attention imbalance.

Current instruction-tuning datasets do not involve joint reasoning over multiple modalities. This is a known problem --- curating data for non-textual modalities is expensive \citep{Liu2023VisualIT, dai2023instructblip}. It is infeasible to curate large amounts of joint reasoning data on top of the instruction data from each modality. However, we hypothesize that correcting for cross-modal attention imbalance is already sufficient to promote joint reasoning. A simple and scalable way to do this is to simply concatenate instructions from multiple modalities within the same context. In other words, we can repurpose existing datasets with this twist to greatly improve cross-modal joint reasoning in \lm{}s.

In summary, we uncover a new fundamental gap between modalities -- in terms of how they are processed in context. We demonstrate that state-of-the-art models fail in a simple cross-modal reasoning task of handling conflicting evidence from multiple modalities. We trace this failure to an imbalance in attention weights across modalities that can be addressed simply by mixing existing instruction data to create cross-modal instructions. %
Our findings also generally highlight the need for training paradigms that mirror the real-world complexity faced by \lm{}s.

\section{Related Work}

\paragraph{Modality Gap in Foundation Models}

\lm{}s are known to fall short when handling inputs from low-resource modalities. For example, some studies demonstrate that \lm{}s answer visual questions primarily based on language priors, disregarding actual visual inputs \citep{Winterbottom2020OnMB,Niu2020CounterfactualVA,Lin2023RevisitingTR}; others illustrate scenarios where models hallucinate objects absent from the image during open-ended generation tasks \citep{Sun2023AligningLM}. Yet, it remains unclear whether this behavior originates primarily from a context-parametric gap \citep{goyal2024context} or a modality gap \citep{Liang2022MindTG}.

Modern vision-language models (VLMs) embed text and images into a shared embedding space \citep{radford2021learning, jia2021scaling}. The modality gap is characterized as separation between the embeddings of different data modalities \citep{Liang2022MindTG} which hurts performance on visual question answering and classification tasks \citep{guo2023modality, winterbottom2020modality}. Several explanations have been proposed, including inductive bias of encoders and disuniformity of contrastive loss \citep{ fahim2024notmodalitygap}. A similar phenomenon persists in multilingual \lm{}s too \citep{Nigatu2023TheLT, Chang2022TheGO}.

\paragraph{Multilingual and Multimodal Instruction Tuning}
Multilingual and vision-language models employ specialized pre-training data \citep{Ustun2024AyaMA, li2024llavanext-strong} and instruction-tuning datasets \citep{li2023bactrianx, Liu2023VisualIT, antol2015vqa} to improve performance on underrepresented modalities. In general, however, these models are designed for unimodal performance and they do not saturate large cross-modal benchmarks like MMMU and ScienceQA that require simultaneously reasoning over data in multiple modalities \citep{yue2023mmmu, lu2022scienceqa}.

Several fine-tuning approaches have been suggested to improve cross-modal reasoning in \lm{}s. For example, X-InstructBLIP claims that training on individual modalities can result in emergent cross-modal reasoning \citep{panagopoulou2023x}. However, we find strong evidence that this is not always possible. 

\paragraph{Related Approaches for Cross-Modal Reasoning}

Another common approach to improve cross-modal performance is to mitigate language bias, or over-dependence on language priors \citep{Niu2020CounterfactualVA}. This approach prevents the model from ignoring images due to parametric knowledge about the question, but does not counteract bias within the context towards textual evidence 
over image evidence.

Other more specialized approaches include aligning individual entities between modalities \citep{lin2022multi} or learning sparse feature representations that rely less on language priors \citep{guo2021adavqa}. In addition, \cite{li2024mosaic} propose a fine-tuning approach that is similar to our instance-level mixing strategy; however, they focus only on textual data, whereas we focus entirely on mixing modalities.

\paragraph{Knowledge Conflicts in Cross-Modal Contexts}
Even in unimodal settings, \lm{}s sometimes fail to identify when they encounter conflicting information \citep{xu2024knowledge, xie2023adaptive}. In these unimodal scenarios (e.g., outdated facts), there is evidence that \lm{}s have the ability to identify when they don't know an answer \citep{kadavath2022language, yin2023large}. In addition, mitigation strategies like prompting \citep{zhou2023context}, pretraining \citep{li2022large}, and reweighting neurons \citep{shi2024ircan} are known to improve detection but remain limited to specific unimodal contexts. Existing instruction-tuning solutions for conflict detection \citep{wang2023resolving} rely heavily on curated conflict-specific datasets. Furthermore, existing benchmarks focus on textual knowledge conflicts \citep{su2024conflictbank}.

Notably, prior work largely overlooks knowledge conflicts between multiple modalities. \cite{liu2024insight} benchmarks cross-modal conflicts, but focuses on context-parametric conflicts between images and the model's pretrained knowledge. \cite{zhu2024unraveling} also studies cross-modal conflicts, but focuses on parametric conflicts between the language model and visual encoder components of a multimodal \lm{}. To the best of our knowledge, knowledge conflicts have not previously been used to study cross-modal reasoning in a controlled setting.

\section{Stress-testing cross-modal reasoning}

\subsection{Formulation}

We study the free-form generation from a \lm{}. The \lm{} takes a context $C$ and a question $Q$ as input and samples a response $y \sim \lm{}(C, Q)$. The context $C$ has an in-context knowledge conflict, i.e., $C$ contains two subsequences, $C_1$ and $C_2$, that support contradictory answers to $Q$. We consider $(C, Q)$ to be a unimodal conflict if $C_{1}, C_{2} \in M_1$ and a cross-modal conflict if $C_1 \in M_1, C_2 \in M_2$ where $M_1, M_2$ are distinct modalities. This allows us to examine whether \lm{}s prioritize one evidence over another or reason jointly to reconcile the conflict. 
Given a set of context-question pairs $\mathcal{D} = \{(C_i, Q_i)\}_{i=1}^N$, we define the \textit{conflict detection rate} as the proportion of samples that are mentioned to contain conflicts.  We used \gptfo as the evaluator (see prompts in \S\ref{app:gpt-evaluator}). 
To isolate the context-based reasoning independent of the parametric bias, we focus on tasks that depend on the context and cannot be solved with the parametric knowledge alone.

\subsection{Data curation}

We construct two datasets: a \textit{cross-modal question answering (CMQA)} dataset and a \textit{cross-lingual question answering (CLQA)} dataset, each with controlled variations in context. See examples in \S\ref{app:dataset-example}.

\paragraph{Cross-modal question answering (CMQA)} The multimodal question answering dataset is constructed over both image and text based on the VQA-v2 dataset \citep{Goyal2016MakingTV}. Each sample in VQA-v2 consists of an image $V$, a question $Q$, and 10 candidate answers. 
In total, we subsample 500 triples of image, question, and answer $(V, Q, A)$ from the dataset. 

For each triplet, we prompt \gptfo{} to generate a text description $\overline{T}$ that does not agree with the image $V$ regarding the question $Q$, and the answer $\overline{A}$ based on $\overline{T}$. Given the image $V$, the text description $\overline{T}$, and the question $Q$, the \lm{} should report a conflict as $A$ is contradictory to $\overline{A}$. We name this dataset $\{(V, \overline{T}, Q, A, \overline{A})\}$ as \texttt{Text-Image}. For each image $V$, we prompt \gptfo{} to generate a description $T'$ that agrees with the image regarding the question $Q$. We name $\{(T', \overline{T}, Q, A, \overline{A})\}$ as \texttt{Text-Text}. For each $\overline{T}$, we prompt \dalle 3 \citep{BetkerImprovingIG} to generate an image $\overline{V}$. We name $\{(V, \overline{V}, Q, A, \overline{A})\}$ as \texttt{Image-Image}.

\paragraph{Cross-lingual question answering (CLQA)} We create a dataset of question answering over synthetic news paragraphs about fictitious events (so the \lm{} cannot use parametric knowledge to answer the questions). We use \gptfo{} to generate 400 topics. For each topic, we prompt \gptfo{} to generate: (1) a synthetic news paragraph $P_E$ in English which has not appeared in reality, a question $Q$ in English, and an answer $A$ based on the paragraph, and (2) synthetic news paragraph $\overline{P}_C$ in Chinese that does not agree with the English one $P_E$ regarding the question $Q$, and an answer $\overline{A}$ based on the Chinese one.

Given the two news paragraphs $P_E$ and $\overline{P}_C$ and the question $Q$, the \lm{} should reason over both since $A$ is contradictory to $\overline{A}$. We name this cross-modal dataset $\{(P_E, \overline{P}_C, Q, A, \overline{A})\}$ as \texttt{English-Chinese}. We then derive several monolingual variants of different language combinations via (back-) translation. For each paragraph $\overline{P}_C$ in Chinese, we back-translate it into English $\overline{P}_E'$. We name $\{(P_E, \overline{P}_E', Q, A, \overline{A})\}$ as \texttt{English-English}. For each English paragraph $P_E$, we translate it into Chinese $P_C'$. We name $\{(P_C', \overline{P}_C, Q, A, \overline{A})\}$ as \texttt{Chinese-Chinese}. Similarly, we test other cross-lingual variants where Chinese is replaced with low-resource languages such as Turkish or Icelandic.

\subsection{Experimental setup}
\label{sec:exp-setup}

We evaluate a range of state-of-the-art (multimodal) \lm{}s on our conflict detection tasks. Multimodal \lm{}s can be applied to both CLQA and CMQA, while text-only \lm{}s can only be applied to CLQA. For text-only \lm{}s, we use Llama-3 and Llama-3.1 \citep{grattafiori2024llama}, Gemma-2 \citep{Riviere2024Gemma2I}, and Aya-23 \citep{Ustun2024AyaMA}.
For multimodal \lm{}s, we use \gptfo{} \citep{gpt4o}, LLaVA-NeXT \citep{li2024llavanext-strong}, and Cambrian \citep{Tong2024Cambrian1AF}. We provide the prompts we use in \S\ref{subsec:prompt-lm}.

\begin{figure}[!th]
\centering
\begin{minipage}{0.58\linewidth}
    \centering
    \includegraphics[width=\linewidth]{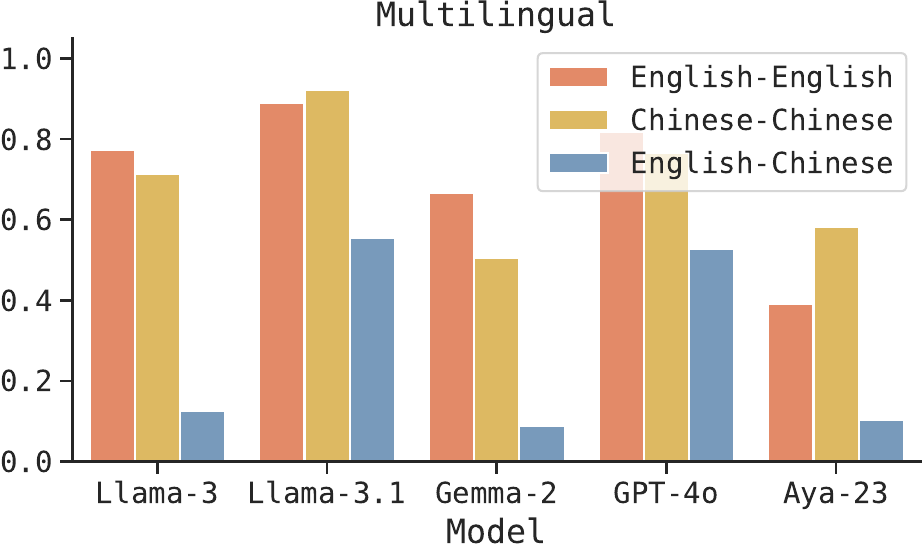}
\end{minipage}\hfill
\begin{minipage}{0.38\linewidth}
    \centering
    \includegraphics[width=\linewidth]{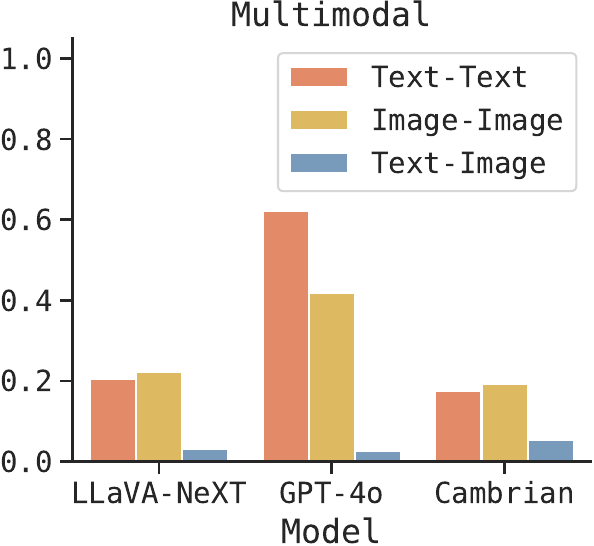}
\end{minipage}
\caption{\label{fig:heterogeneity} \lm{}s are worse at reasoning over cross-modal contexts than unimodal contexts.}
\end{figure}

\subsection{Results}
\label{subsec:result}

Figure~\ref{fig:heterogeneity} shows the performance of the \lm{}s on our CLQA and CMQA datasets. We see that the conflict detection performance is significantly lower with the cross-modal contexts than with the unimodal contexts.

For CLQA (Figure~\ref{fig:heterogeneity} left), we the performance on \texttt{English-English} is comparable to \texttt{Chinese-Chinese}, and both are far better -- up to 5x -- than \texttt{English-Chinese}. This shows that the lower performance in the multilingual setting is not due to the limited general capability of the \lm{} in Chinese. Also, recall that the questions are always in English, including in the \texttt{Chinese-Chinese} setting, so the lower performance with multilingual contexts is neither due to the language barrier between English and Chinese. Experimental results for Turkish and Icelandic are similar to those for Chinese, so we put them in \S\ref{app:languages} for conciseness. 
We see similar trends on the CMQA task (Figure~\ref{fig:heterogeneity} bottom) -- the performance with unimodal contexts (\texttt{Text-Text} and \texttt{Image-Image}) is far better than the performance with cross-modal contexts (\texttt{Text-Image}) for all \lm{}s.

\begin{figure}[!th]
\centering
\begin{minipage}{0.4\linewidth}
    \centering
    \includegraphics[width=\linewidth]{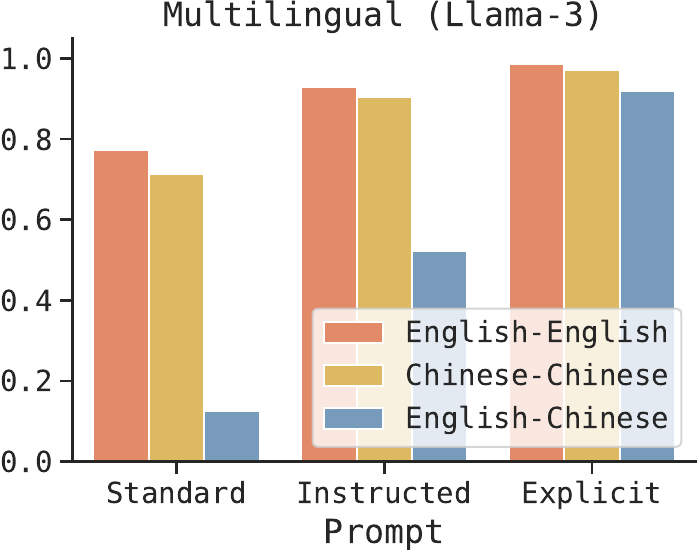}
\end{minipage} \hspace{15pt}
\begin{minipage}{0.4\linewidth}
    \centering
    \includegraphics[width=\linewidth]{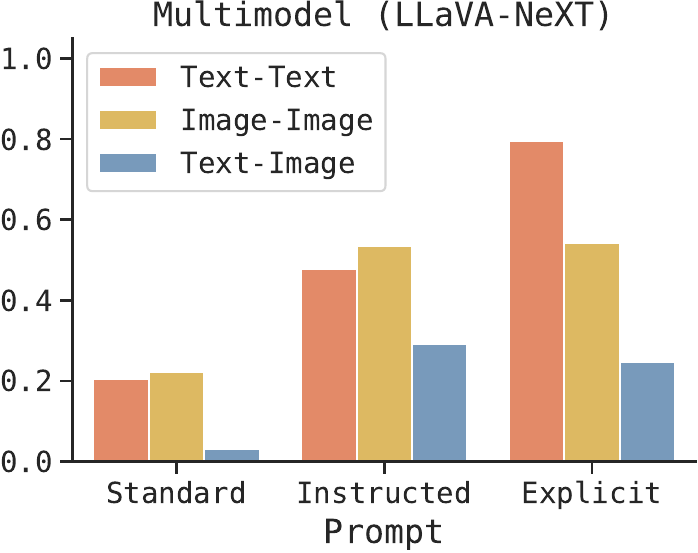}
\end{minipage}
\caption{\label{fig:prompt} Ablation studies on the prompt. \lm{}s are worse at reasoning over cross-modal contexts than unimodal contexts. 
See the text for details of each prompt.
}
\end{figure}

In addition to the prompts above (denoted as \texttt{Standard} in Figure~\ref{fig:prompt}), which do not assume intervention from the user, we also test other prompts that encourage \lm{} to detect the conflict. Specifically, we explore two types of prompts: (1) add an instruction that tells the \lm{} to report the conflict if it finds any (denoted as \texttt{Instructed} in Figure~\ref{fig:prompt}); (2) embed the question into a yes-no question: ``Would the answers to the question `$\{Q\}$' be the same based on the paragraphs in the context?'' (denoted as \texttt{Explicit} in Figure~\ref{fig:prompt}). In Figure~\ref{fig:prompt}, we see that, although the overall conflict detection performance improves, the trend is similar to Figure~\ref{fig:heterogeneity} -- the conflict detection performance is lower in the cross-modal contexts than in the unimodal contexts. In the next section, we explore why this is the case and try to improve this practically.

{\centering
\begin{tcolorbox}[left=1mm, right=1mm, width=\linewidth, colback=blue!5!white,colframe=white]
\textbf{Summary.} State-of-the-art \lm{}s fail in a simple cross-modal reasoning task of handling conflicting evidences in multiple modalities.
\end{tcolorbox}
}

\section{Cross-modal attention imbalance}

\subsection{Attention imbalance in \lm{}s}
\label{sec:attention-imbalance}

To investigate the mechanisms underlying this failure, we probe the context contribution in \lm{}s. Most state-of-the-art \lm{}s are autoregressive -- at each step, the \lm{} predicts the next token based on the context so far.
For architectures like Transformers \citep{Vaswani2017AttentionIA}, the representation at each step can be decomposed into a linear combination of the contributions of each span of context. For example, in a Transformer \lm{}, the output of an attention head in a layer at step $t$ is defined as:
\begin{align}
    \bm{a}_t = \bm{W}_O \sum_{j=1}^{t} w_{t, j} \bm{v}_j ,
\end{align}
where $\bm{v}_j$ is value output of the $j$-th token in the context, $w_{t, j}$ is the attention weight from the $t$-th token to the $j$-th token, and $\bm{W}_O$ is the output projection matrix. We can group tokens in the context based on their domain: $\mathcal{C}_k$ contains all token indices of the $k$-th group. Based on this, we can rewrite the equation as:
\begin{align}
\label{eq:attention-decomposition}
    \bm{a}_t = \sum_{k=1}^{K} \left( \sum_{j \in \mathcal{C}_k} w_{t, j} \bm{W}_O\bm{v}_j \right) := \sum_{k=1}^{K}\bm{u}_k.
\end{align}
The term $\bm{u}_k$ is a vector that the $k$-th context writes to the residual stream at step $t$. It shows that the context representation is a linear combination of each context's contribution. 

We hypothesize that the context contribution \textit{in the task-relevant subspaces} is imbalanced in the cross-modal contexts, making the \lm{} more likely to rely on the dominant context instead of doing conflict detection. In Figure~\ref{fig:illustration}, we illustrate our mental model of attention imbalance. In unimodal contexts (A), different domains show balanced normalized attention ($\mathrm{softmax}(QK^\top)$) despite divergent pre-softmax logits ($QK^\top$). Cross-modal contexts (B) expose cross-modal attention imbalance -- normalization fails to mitigate logit-level imbalance. Instance-level modality mixing (C) resolves this by training models to intrinsically balance attention logits across co-occurring domains.

\begin{wrapfigure}[16]{r}{0.55\textwidth}
\vspace{-10pt}
\centering
    \includegraphics[width=0.48\linewidth]{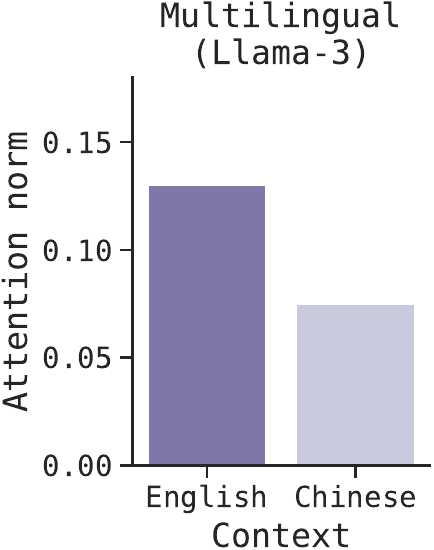} \hfill
    \includegraphics[width=0.48\linewidth]{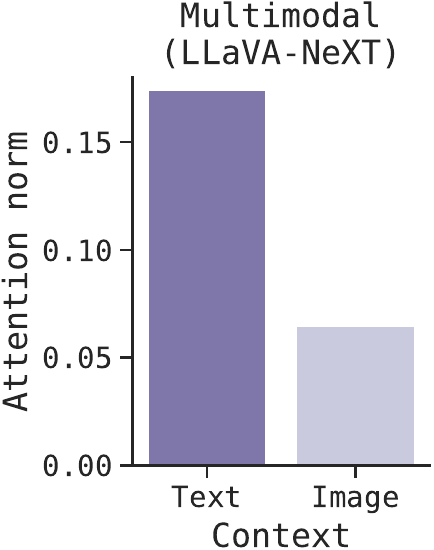}
    \caption{\label{fig:attn_plot} Cross-modal attention imbalance. English has larger attention than Chinese and images.}
\end{wrapfigure}

To demonstrate attention imbalance, we compute the average norm of $\bm{u}_k$ for each context, averaged over all layers and attention heads.\footnote{We note that $\bm{u}_k$ averaged over all layers and attention heads should be viewed as a proxy of what we want to measure, i.e., context contribution \textit{in the task-relevant subspaces}. We further discuss this in \S\ref{app:attention-visualization}. For this reason, we do not argue that the norms of different $\bm{u}_k$ should be the same to achieve the best conflict detection performance.} Figure~\ref{fig:attn_plot} shows that, for cross-lingual, the English context contributes more than the Chinese context; for cross-modal, text contributes more than images.

To test if there is a \textit{causal} relationship between attention imbalance and cross-modal reasoning, we causally intervene the contribution of a context $\mathcal{C}_k$ by adding a small constant $\epsilon$ to its unnormalized attention score. Formally, denote the normalized attention weights at step $t$ as $\bm{w}_t := [w_{t, 1}, \ldots, w_{t, t}]^\top$. We  manipulate the attention weights as follows:

$\label{eq:manip}
    \mathrm{Manip}(\bm{w}_t) = \mathrm{softmax}\left(\log\bm{w}_t + \epsilon \mathbf{1}_{\mathcal{C}_k}\right),$

where $\mathbf{1}_{\mathcal{C}_k}$ is a vector with $1$'s on all the $\mathcal{C}_k$ context positions and $0$'s otherwise. 

\begin{figure}[!th]
\centering
    \includegraphics[width=0.28\linewidth]{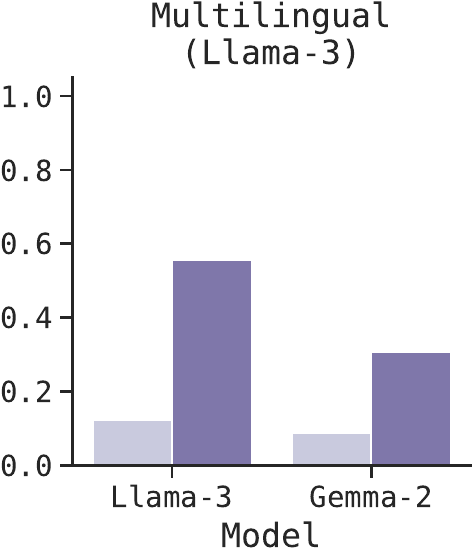}\hspace{0.02\linewidth}
    \includegraphics[width=0.28\linewidth]{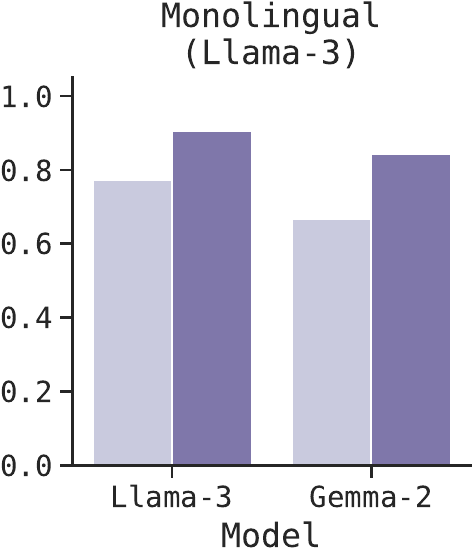}\hspace{0.02\linewidth}
    \includegraphics[width=0.17\linewidth]{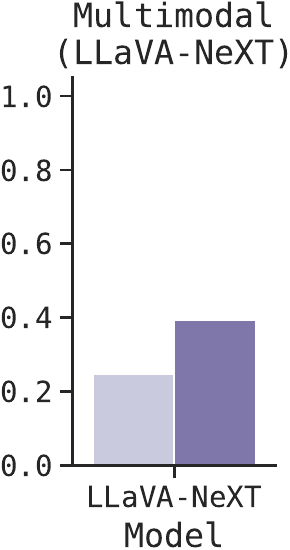} \hspace{0.01\linewidth}
    \raisebox{55pt}[0pt][0pt]{%
        \includegraphics[width=0.17\linewidth]{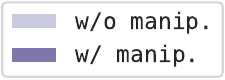}
    } 
    \caption{\label{fig:manipulate} Cross-modal attention imbalance has a causal effect on cross-modal reasoning: we apply a fixed attention bias to increase the attention over the context with a smaller attention output norm, and see that this improves the conflict detection performance. We use the \texttt{Standard} prompt for cross-lingual and monolingual settings, and \texttt{Explicit} prompt for the cross-modal setting. %
    }
\end{figure}

Figure~\ref{fig:manipulate} shows that the conflict detection performance indeed improves after attention manipulation. In the cross-lingual setting, the absolute improvement is up to 43\% (relative by 5x). In the cross-modal setting, we observe a smaller yet significant gain of 18\%. 
As a side observation, we find that attention manipulation can help the unimodal context as well: we find that \lm{}s exhibit primacy bias and tend to rely more on the context that appears first. By increasing the attention weights on later context, we also further improve the conflict detection performance on unimodal context. %

\subsection{Verification on other datasets}  
\label{verification-other-datasets}

To verify the generality of our findings, we applied the attention manipulation to three benchmarks that require substantive visual reasoning. Popular datasets such as MMMU show only marginal performance differences between text‑only and vision-language models, which limits their diagnostic value \citep{Fu2024BLINKML}.
In contrast, we use two more challenging datasets: BLINK \citep{Fu2024BLINKML} and SAT \citep{Ray2024SATDS}.

\begin{wrapfigure}[11]{r}{0.6\textwidth}
\centering   \includegraphics[width=\linewidth]{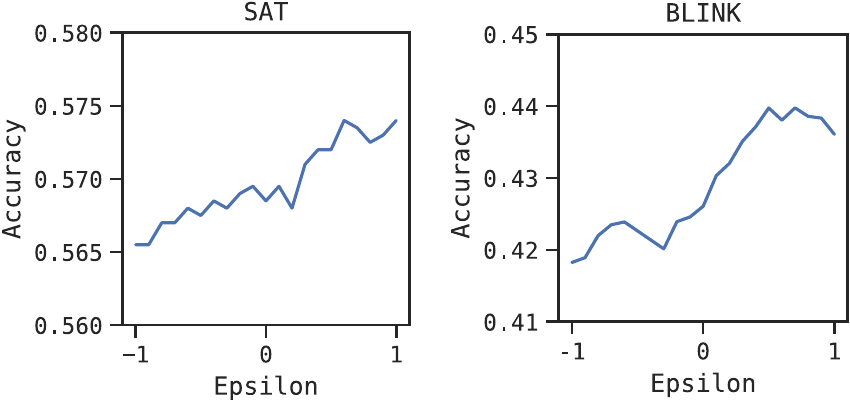}
    \caption{\label{fig:attention-on-others} The performance on cross-modal reasoning datasets improves as we increase the strength of attention manipulation. %
    }
\end{wrapfigure}
We apply the attention manipulation to the LLaVA-NeXT 8B model, where we sweep the manipulation strength from -1 (less visual attention) to 1 (more visual attention). 
Figure~\ref{fig:attention-on-others} shows that attention manipulation achieves an absolute accuracy gain of 1\% and 2\%. These results indicate that reducing attention imbalance between modalities is especially effective when visual reasoning is indispensable.

{\centering
\begin{tcolorbox}[left=1mm, right=1mm, width=\linewidth, colback=blue!5!white,colframe=white]
\textbf{Summary.} Cross-modal attention imbalance has a causal negative effect on \lm{}s' cross-modal reasoning capability.
\end{tcolorbox}
}

\section{How does training affect cross-modal attention imbalance?}

\subsection{Dataset-level modality mixing does not help}
\label{subsec:dataset-level-mixing-fails}

We begin by noting that most state-of-the-art \lm{}s today are trained on highly diverse corpora, spanning a wide range of domains and multiple languages \citep{grattafiori2024llama,Riviere2024Gemma2I}. More surprisingly, as we observe in Figure~\ref{fig:heterogeneity}, Aya-23, a \lm{} specifically optimized for multilingual capabilities, performs no better than other \lm{}s with multilingual contexts. This suggests that simply training \lm{}s on diverse modalities does not, by itself, ensure good cross-modal reasoning.

\begin{figure}[!th]
\centering
    \includegraphics[width=0.3\linewidth]{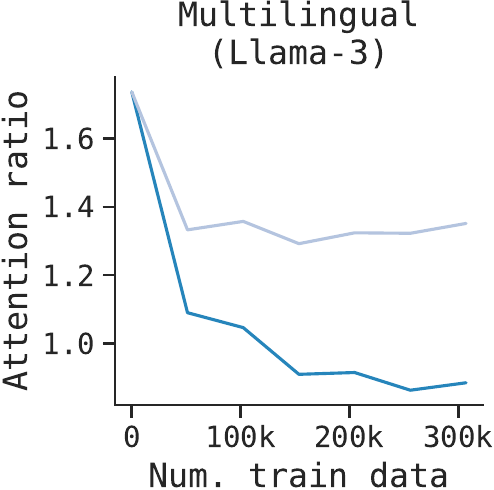} 
    \includegraphics[width=0.3\linewidth]{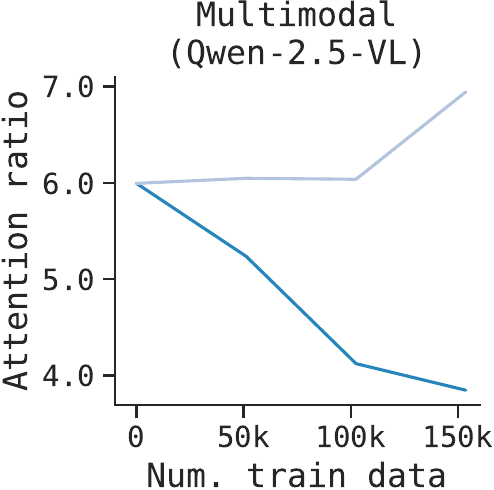}
    \raisebox{45pt}[0pt][0pt]{%
        \includegraphics[width=0.35\linewidth]{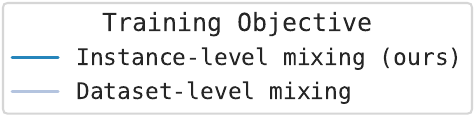}
    }
    \caption{\label{fig:finetuning-attention} Finetuning on instance-level mixed data (dark blue) reduces cross-modal attention imbalance significantly more than fine-tuning on dataset-level mixed data (gray).
    }
\end{figure}

To reinforce this, we run two instruction-tuning experiments. First, we finetune Llama-3 on mixed English and Chinese instruction tuning datasets (we call this strategy \textbf{dataset-level modality mixing}) and see if this improves conflict detection in the cross-lingual \texttt{English-Chinese} setting. Specifically, we use the English and Chinese subsets of Bactrian-X \citep{li2023bactrianx}, a multilingual instruction-tuning dataset containing 67k samples in each language. Similarly, we finetune Qwen-2.5-VL on mixed text and visual instruction tuning datasets (\textbf{dataset-level modality mixing}) and see if this improves conflict detection in the cross-modal \texttt{text-image} setting. In this experiment, we use the visual instruction data from \citet{liu2024visual} and the English subset of Bactrian-X. In Figure~\ref{fig:finetuning-attention}, we see that \textit{dataset-level modality mixing} offers minimal gains in alleviating cross-domain attention imbalance. This motivates us to understand why diverse, multimodal data is not enough to close the gap between unimodal and cross-modal contexts. 

\subsection{Instance-level modality mixing}

We have shown that standard cross-modal instruction tuning (e.g. having both English and Chinese examples in the data, or having both text and visual instruction tuning examples) fails to improve cross-modal attention imbalance.
We hypothesize that the gap in unimodal and cross-modal contexts arises because mixing datasets does not expose models to instances requiring cross-domain reasoning \textit{within the same context}. Without instance-level modality mixing between modalities, the pre-softmax attention scores for one domain could be hugely different from that of another domain, without changing the normalized attention scores on each domain (Figure~\ref{fig:illustration}). To address the lack of instance-level modality mixing between modalities, we propose a simple and scalable method of explicitly combining multiple modalities within each training instance. Here is an illustration of our input and output format in the cross-lingual setting:

{\centering
\begin{tcolorbox}[left=1mm, right=1mm, width=\linewidth, colback=orange!6!white,colframe=white]
    \textit{Input:} \\
    \texttt{<Chinese instruction> <English instruction> \\
    Reply to both user instructions.} \\
    \textit{Output:} \\
    \texttt{<Chinese response> <English response>}
\end{tcolorbox}
}

In the cross-modal setting we have the input and output as:

{\centering
\begin{tcolorbox}[left=1mm, right=1mm, width=\linewidth, colback=orange!6!white,colframe=white]
    \textit{Input:} \\
    \texttt{<text instruction> \\ <image> <image-related instruction> \\
    Reply to both user instructions.} \\
    \textit{Output:} \\
    \texttt{<text response> <image-related response>}
\end{tcolorbox}
}

\begin{figure}[!th]
\centering
    \includegraphics[width=0.3\linewidth]{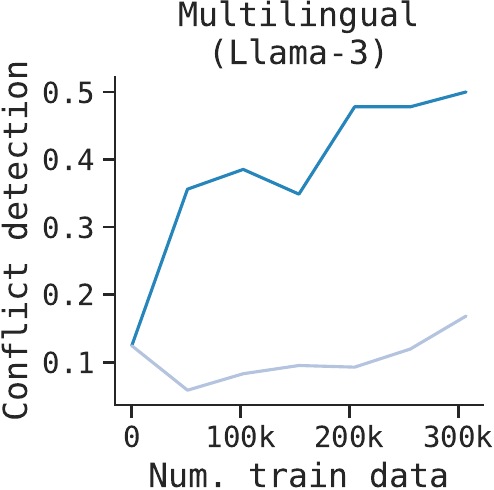}
    \includegraphics[width=0.3\linewidth]{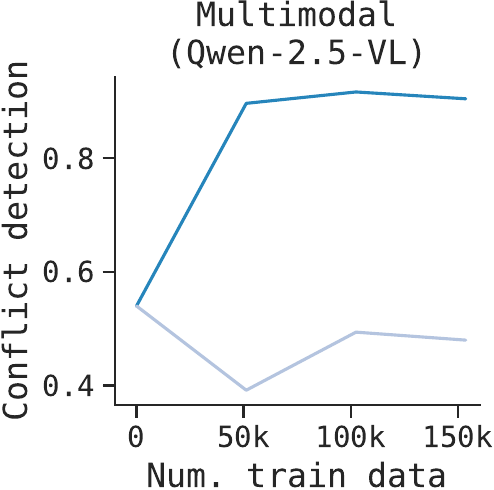}
    \raisebox{45pt}[0pt][0pt]{%
        \includegraphics[width=0.35\linewidth]{figures/legend.pdf}
    }
    \caption{\label{fig:finetuning-performance} Finetuning on instance-level mixed data (dark blue) improves cross-modal conflict detection largely more than fine-tuning on traditional dataset-level mixed data (gray).
    }
\end{figure}

To verify the benefit of instance-level modality mixing, we use the same data from \S~\ref{subsec:dataset-level-mixing-fails} but mix them at the instance level instead of at the dataset level. In Figure~\ref{fig:finetuning-attention}, we report the attention imbalance of model checkpoints for instance-level modality mixing and the dataset-level modality mixing baseline in \S\ref{subsec:dataset-level-mixing-fails}. In the cross-lingual setting, instance-level modality mixing reduces attention imbalance between modalities by $4\times$. In the cross-modal setting, it reduces attention imbalance by $34\%$. In Figure~\ref{fig:finetuning-performance}, we report the performance of model checkpoints for \textit{instance-level modality mixing} and the baseline in \S\ref{subsec:dataset-level-mixing-fails} (\textit{dataset-level modality mixing}). In the cross-lingual setting, instance-level modality mixing boosts conflict detection by $37\%$, much greater than dataset-level modality mixing. In the cross-modal setting, instance-level modality mixing improves conflict detection by $2\times$.

We highlight that instance-level modality mixing is more scalable than directly finetuning the \lm{}s on the knowledge conflict detection task itself, as it does \emph{not} require any explicit conflicts within the instructions, which could be costly to generate for diverse domains. Notably, the improvement in conflict detection does not come from training on the same task that we are testing on, but rather a general proxy for cross-modal attention balance.

{\centering
\begin{tcolorbox}[left=1mm, right=1mm, width=\linewidth, colback=blue!5!white,colframe=white]
\textbf{Summary.} Instance-level modality mixing mitigates attention imbalance and improves cross-modal reasoning, without requiring any additional data curation.
\end{tcolorbox}
}

\subsection{Verifying instance-level modality mixing on downstream tasks}

\begin{wrapfigure}[16]{r}{0.48\textwidth}
\vspace{-10pt}
\centering   \includegraphics[width=\linewidth]{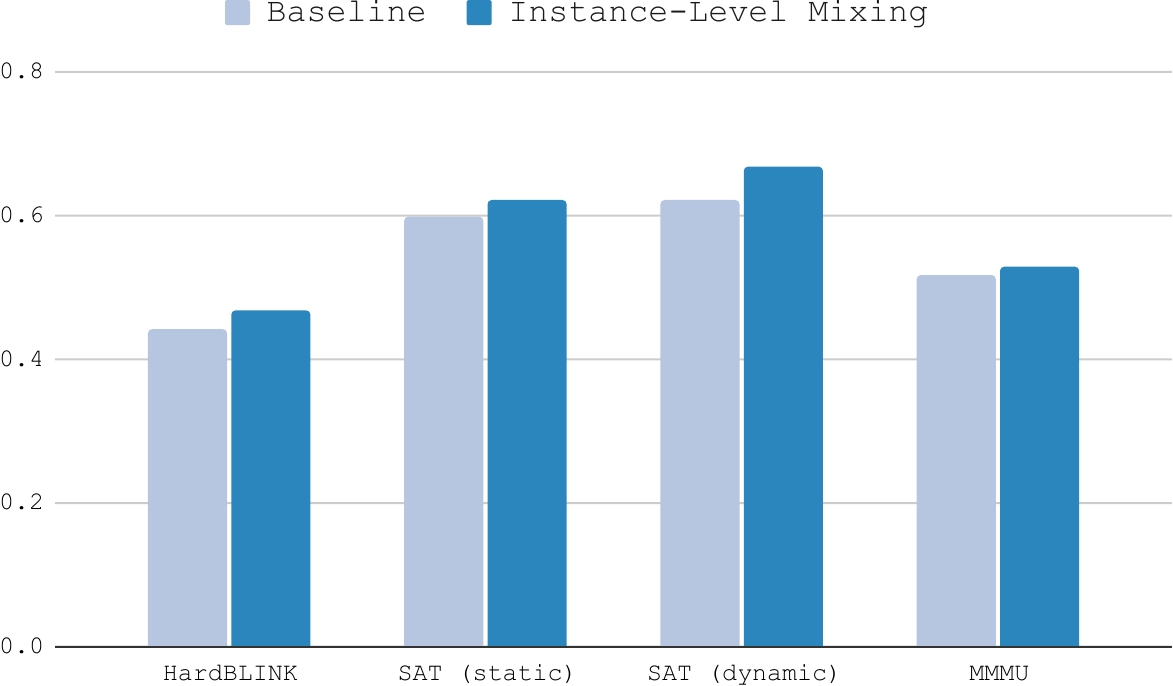}
    \caption{Finetuning Qwen-2.5-VL on instance-level mixed data improves performance on hard multimodal benchmarks.}
    \label{fig:instance-level-downstream} 
\end{wrapfigure}

Synthetic data must be validated against real‑world complexity. Similar to manual attention rebalancing (\S\ref{verification-other-datasets}), we find that instance-level modality mixing improves performance on several hard vision-language benchmarks. These experiments indicate that our proposed methods to improve conflict detection also enhance the overall reasoning capabilities of models, especially on datasets that require non-trivial reasoning over each domain. As demonstrated in Figure \ref{fig:instance-level-downstream}, fine-tuning on instance-level mixed data improves performance by 2.4\% on HardBLINK, 2.4\% on the static subset of SAT, 4.4\% on the dynamic subset of SAT, and 1.2\% on MMMU. Future work could explore additional data curation and instance-level mixing strategies to improve downstream performance gains.

\section{Conclusions}

We uncovered a new fundamental gap in how \lm{}s process modalities in context. Through controlled datasets and experiments, we demonstrated that state-of-the-art models fail in a simple cross-modal reasoning task of handling conflicting evidences in multiple modalities. Our analyses trace the problem to \emph{cross-modal attention imbalance}, an imbalance in attention weights across modalities. We showed that simply including multiple modalities in training (i.e., dataset-level modality mixing) has little gains, while explicitly mixing different modalities within each training sample (i.e., instance-level modality mixing) mitigates attention imbalance and substantially boosts conflict detection. Our results highlight the need for training paradigms that mirror the real-world complexity faced by models and for scaling methods that enable foundation models to balance cross-modal attention and reason on cross-modal contexts.

\section{Impact Statements}

Our findings highlight an important gap in \lm{}'s ability to detect contradictory information when it appears in different modalities or languages. As such models increasingly support real-world applications, especially in high-stake applications like computer-use agents and medical care, overlooking factual conflicts can harm user trust, spread misinformation, and compromise safety. 

This work also provides practical guidance for developers seeking to build reliable AI systems, including post hoc methods and recommendations about training data curation. We hope to contribute to broader efforts aiming to ensure the safety and responsibility of \lm{}s in increasingly complex and diverse applications.

\section*{Acknowledgments}

We express our gratitude to Sachin Goyal and Taeyoun Kim for their valuable feedback on the draft of this paper.

\bibliography{colm2025_conference}
\bibliographystyle{colm2025_conference}

\appendix
\onecolumn
\newpage
\appendix

\section{Additional examples of conflict detection}
\label{app:examples-of-model-failure}

We provide additional examples of \lm{} failure in conflict detection over cross-modal contexts.

\begin{figure*}[!th]
\centering
    \includegraphics[width=1\linewidth]{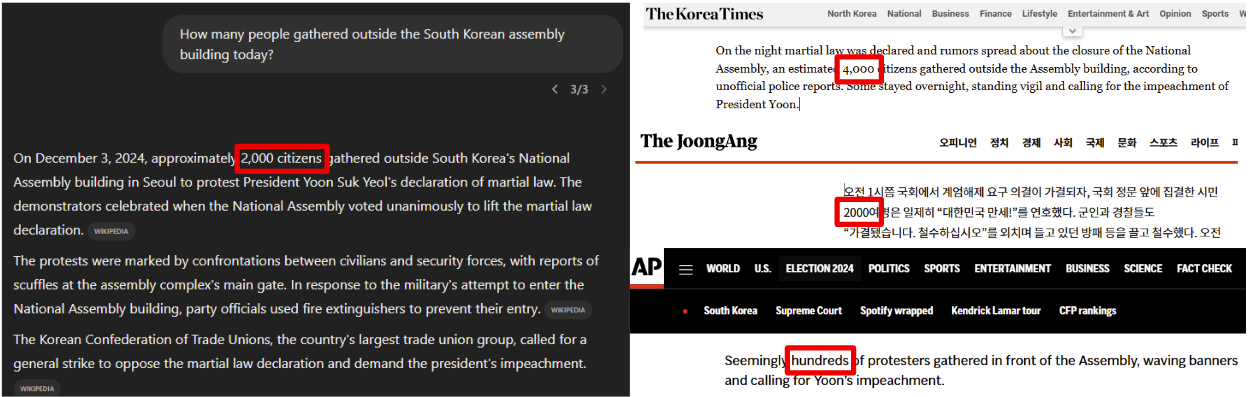}
    \caption{\label{fig:motivation2} An \lm{} with web access can fail to acknowledge knowledge conflicts from multilingual news sources. For example, GPT-4o reports the size of the protest outside South Korea's National Assembly on December 3, 2024 as 2,000 people, although different sources provide conflicting numbers of attendees.}
\end{figure*}

\paragraph{Multilingual example} In Figure~\ref{fig:motivation2}, we provide an example of how an \lm{} (GPT-4o) with web access can fail to acknowledge knowledge conflicts from multilingual news sources. 

\begin{figure*}[!th]
\centering
    \includegraphics[width=1\linewidth]{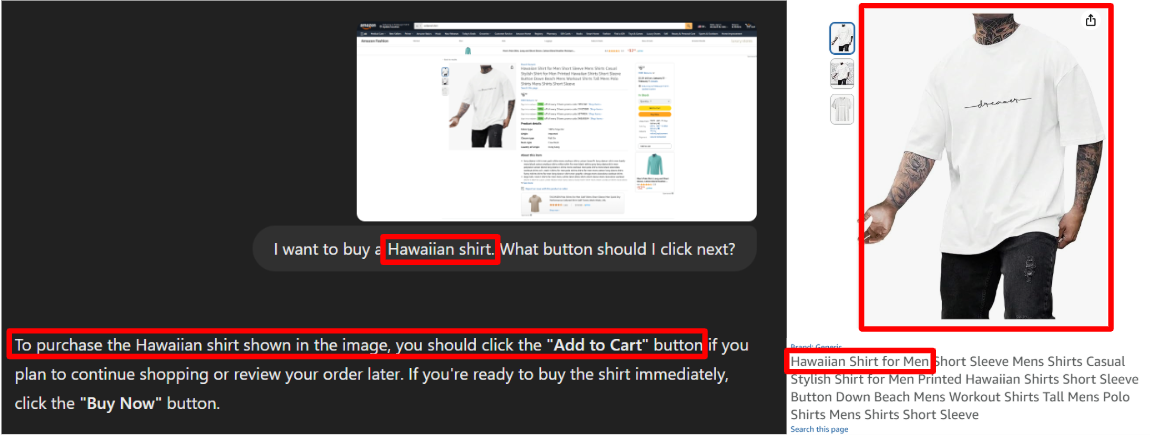}
    \caption{\label{fig:motivation3} A \lm{} can fail to acknowledge knowledge conflicts in multiple modalities. For example, GPT-4o instructs the user to purchase an item labeled as "Hawaiian Shirt for Men" despite the image clearly depicting an ordinary t-shirt, not a Hawaiian shirt.}
\end{figure*}

\paragraph{Cross-modal agent example} In Figure~\ref{fig:motivation3}, we provide an example of how an \lm{} (GPT-4o) with web access can fail to acknowledge knowledge conflicts in multimodal product descriptions.

\newpage
\section{Dataset Examples}
\label{app:dataset-example}

\begin{figure*}[!th]
\centering
    \includegraphics[width=0.8\linewidth]{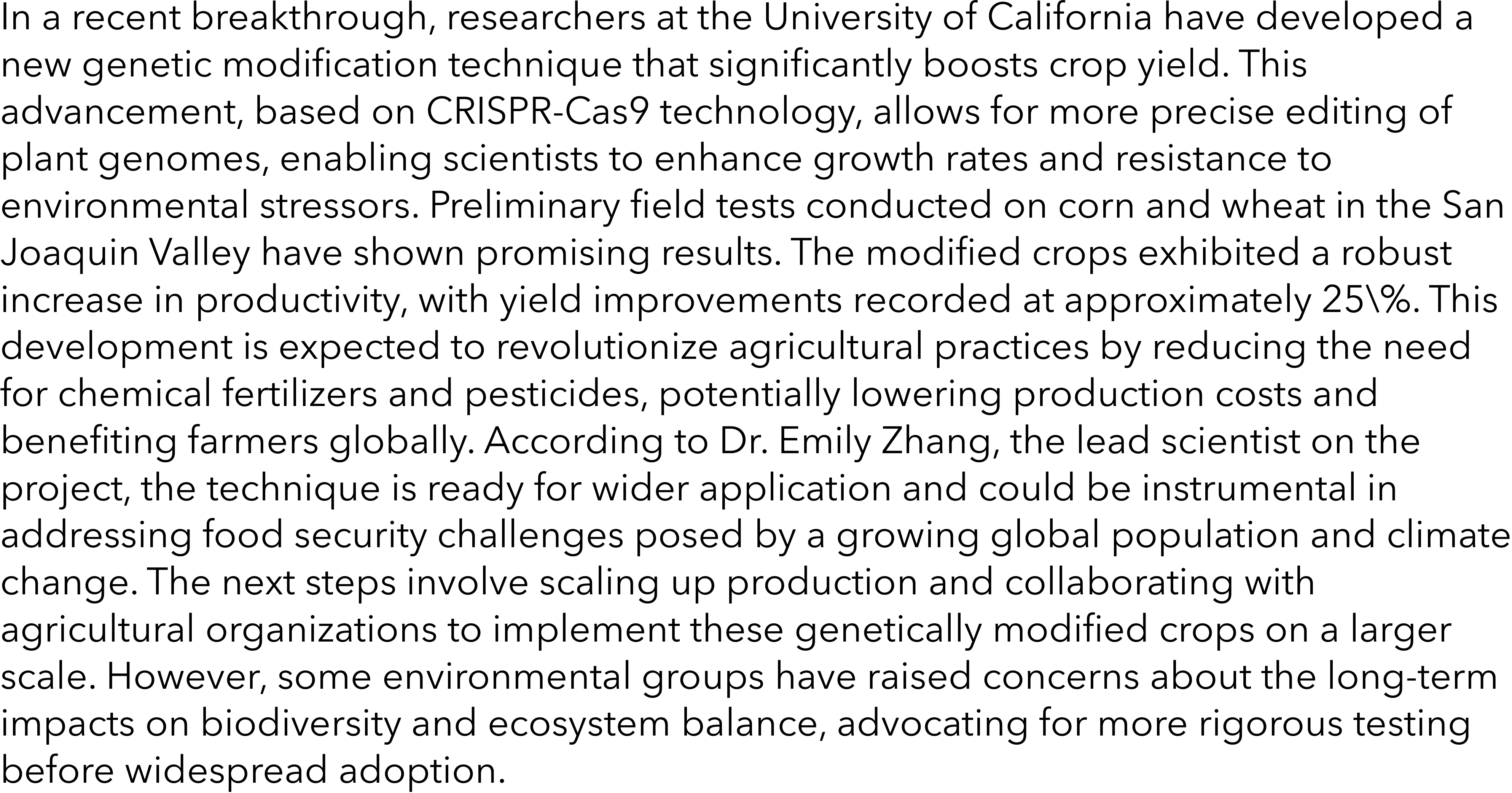} \\ \vspace{5pt}
    \includegraphics[width=0.8\linewidth]{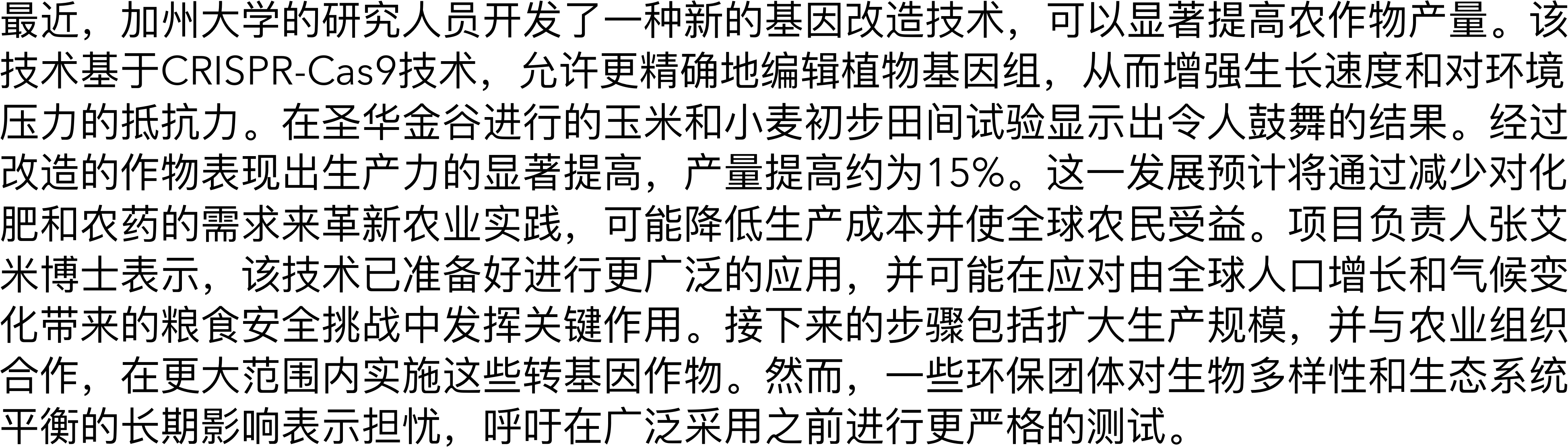} \\ \vspace{5pt}
    \includegraphics[width=0.8\linewidth]{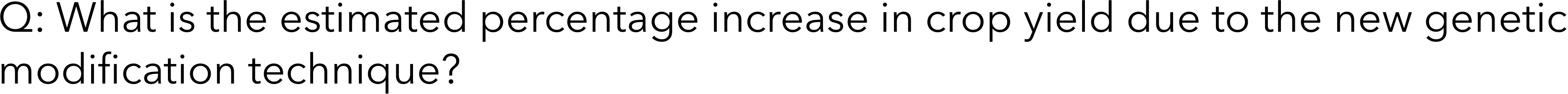}
    \caption{\label{fig:multilingual-example-data} English evidence, Chinese evidence, and question from our CLQA dataset. }
\end{figure*}

\begin{figure*}[!th]
\centering
    \includegraphics[width=0.8\linewidth]{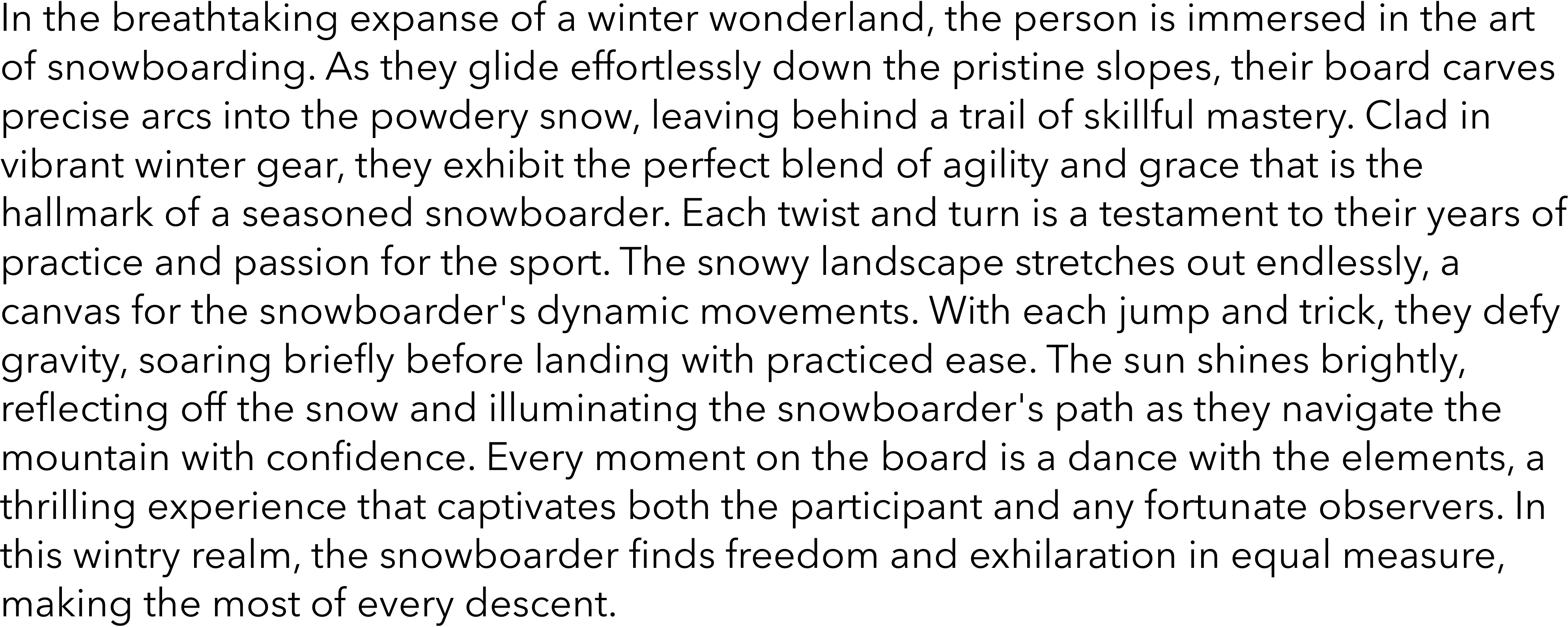} \\ \vspace{5pt}
    \includegraphics[width=0.37\linewidth]{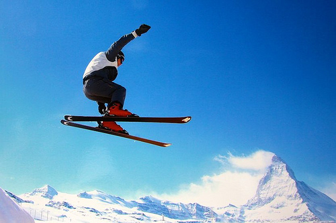} \\ \vspace{5pt}
    \includegraphics[width=0.26\linewidth]{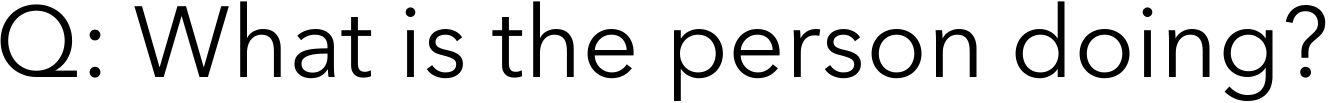}
    \caption{\label{fig:multimodal-example-data} Text evidence, image evidence, and question from our CMQA dataset. }
\end{figure*}

\begin{figure*}[!th]
\centering
    \includegraphics[width=0.37\linewidth]{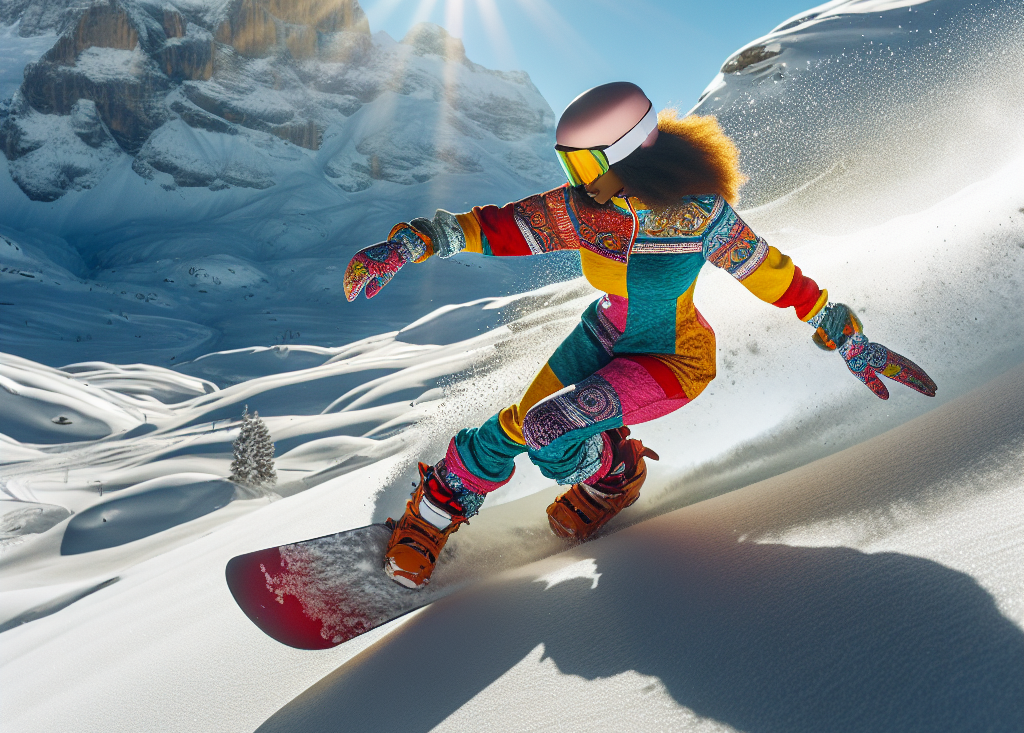} \\ \vspace{5pt}
    \includegraphics[width=0.37\linewidth]{figures/example-image.png} \\ \vspace{5pt}
    \includegraphics[width=0.26\linewidth]{figures/example-q-multimodal.pdf}
    \caption{\label{fig:multimodal-example-data-2} Multi-image evidence and question from our CMQA dataset.}
\end{figure*}

Figure~\ref{fig:multilingual-example-data} shows an example English evidence, Chinese evidence, and question from our CLQA dataset. Figure~\ref{fig:multimodal-example-data} shows an example text evidence, image evidence, and question from our CMQA dataset. Figure~\ref{fig:multimodal-example-data-2} shows an example of two image evidences and question from our CMQA dataset. 

\newpage
\section{Prompts}

\subsection{LM-as-a-Judge for Evaluating Conflict Detection}
\label{app:gpt-evaluator}

We use the following prompt for \gptfo to evaluate if the model output is doing conflict detection.

{\centering
\begin{tcolorbox}[left=1mm, right=1mm, width=\linewidth, colback=red!6!white,colframe=white]
\begin{verbatim}
I'll provide you with a question and a response by a model.

## Your task

Can you infer from the response that the pieces of evidence provided to the model 
has contradictions within them (or that different pieces of evidence suggest 
different answers)? 
If yes, then contradict_score = 1, otherwise contradict_score = 0.

Examples of contradictions: 
- two images describing different things, e.g., one image describes something, 
while the other image describes something else;
- image and text describing different things, e.g., the image describes something, 
while the text describes something else;
- two paragraphs giving different answers, e.g., one paragraph says something, 
while the other paragraph says something else.

## Input

Question: {question}
Response: {response}

## Output requirements

Wrap your final judgement in triple backticks. Your judgement should look like 
this:

```json
{{
    "contradict_score": 0 or 1
}}
```
\end{verbatim}
\end{tcolorbox}
}

\subsection{CMQA and CLQA Prompts}
\label{subsec:prompt-lm}

We prompt \lm{}s with two pieces of context and a question, and sample a response from the \lm{}. Here is the prompt we use for our CLQA task:

{\centering
\begin{tcolorbox}[left=1mm, right=1mm, width=\linewidth, colback=orange!6!white,colframe=white]
    \texttt{Here are some paragraphs.} \\
    \texttt{<paragraph\_1>} \\
    \texttt{<paragraph\_2>} \\
    \texttt{Based on all the paragraphs, answer the question below. Reply in English.} \\
    \texttt{<question>}
\end{tcolorbox}
}

Similarly, we use the following prompt for our CMQA task:

{\centering
\begin{tcolorbox}[left=1mm, right=1mm, width=\linewidth, colback=orange!6!white,colframe=white]
    \texttt{<image>} \texttt{<text>} \quad \textit{// The order depends on the model.} \\
    \texttt{Above are visual and textual descriptions of a scene. \\ Answer the question below.} \\
    \texttt{<question>}
\end{tcolorbox}
}

\section{Additional experiments in attention imbalance}
\label{app:attention-visualization}

Recall that in \S\ref{sec:attention-imbalance} we demonstrate cross-modal attention imbalance with the average norm of $\bm{u}_k$ for each context, averaged over all layers and attention heads. In this section, we elaborate on this by visualizing $\bm{u}_k$ for each context in each layer and attention head. 

\begin{figure*}[!th]
\centering
    \includegraphics[width=\linewidth]{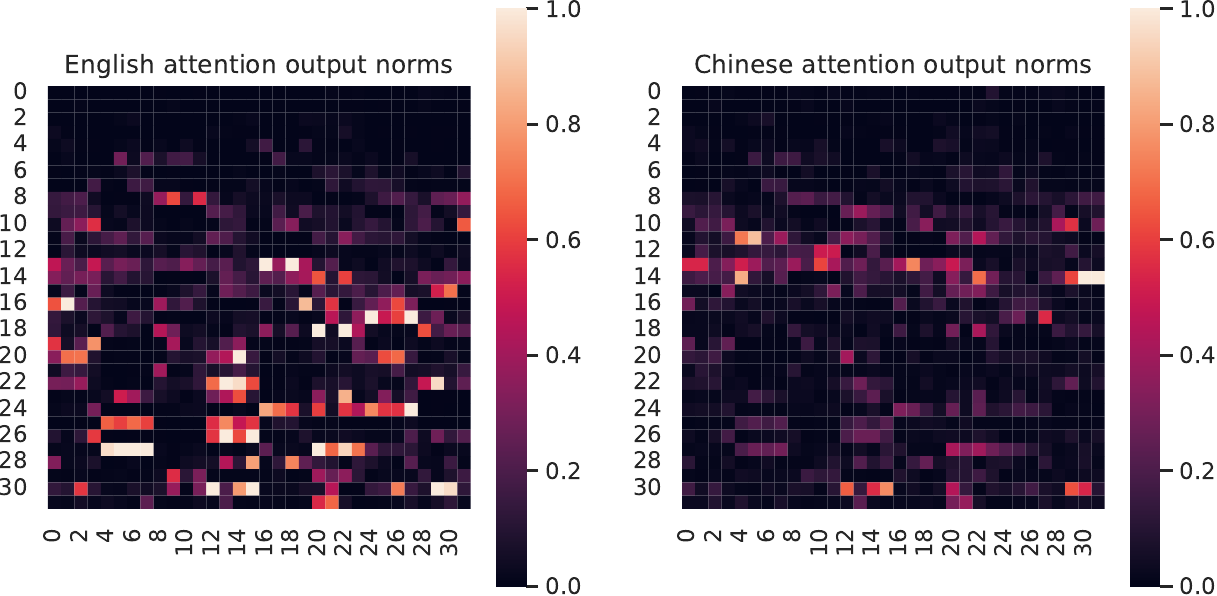}
    \caption{\label{fig:attention-visualization-multilingual} We visualized the norm of $\bm{u}_k$ for each layer and attention head in the multilingual setting, aggregated over all test samples. We see that the values over the Chinese context is generally smaller than those over the English context, especially in upper layers. Notably, Layers 11-14 are an exception, where the values over the Chinese context is higher -- we argue that these layers are \textit{not} in the task-relevant subspace (i.e., they activates on the Chinese context but does not improve the reliance on Chinese when answering the question).}
\end{figure*}

\begin{figure*}[!th]
\centering
    \includegraphics[width=\linewidth]{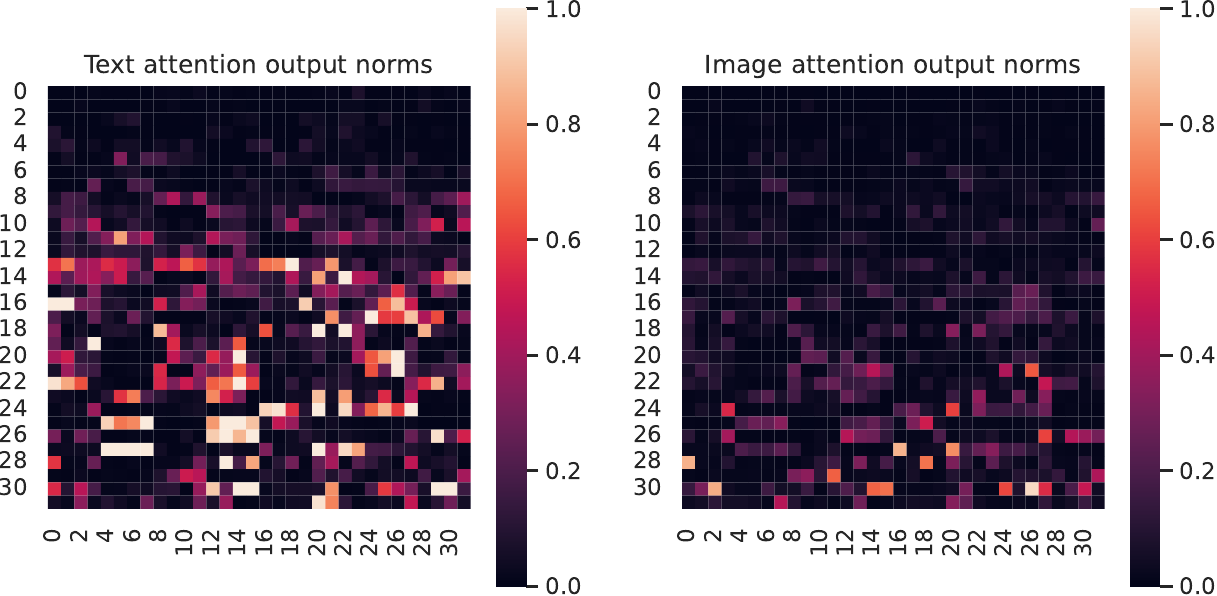}
    \caption{\label{fig:attention-visualization-multimodal} We visualized the norm of $\bm{u}_k$ for each layer and attention head in the multimodal setting, aggregated over all test samples. We see that the values over the image is generally smaller than those over the text.}
\end{figure*}

Figure~\ref{fig:attention-visualization-multilingual} visualizes the norm of $\bm{u}_k$ for each layer and attention head in the multilingual setting, aggregated over all test samples. Figure~\ref{fig:attention-visualization-multimodal} visualizes the norm of $\bm{u}_k$ for each layer and attention head in the multilingual setting, aggregated over all test samples. We see that that the values over the Chinese/image context is generally smaller than those over the English/text context, especially in upper layers. 

We note one important exception that is relevant to the footnote in \S\ref{sec:attention-imbalance}, where we argue that $\bm{u}_k$ averaged over all layers and attention heads should be viewed as a proxy of what we want to measure, i.e., context contribution \textit{in the task-relevant subspaces}. Figure~\ref{fig:attention-visualization-multilingual}, Layers 11-14 are an exception, where the values over the Chinese context is higher -- we argue that these layers are \textit{not} in the task-relevant subspace (i.e., they activates on the Chinese context but does not improve the reliance on Chinese when answering the question). For this reason, we do not argue that the norms of different $\bm{u}_k$ should be the same to achieve the best conflict detection performance.

\section{Additional results on other languages}
\label{app:languages}

\begin{figure}[!th]
\centering
    \includegraphics[width=\linewidth]{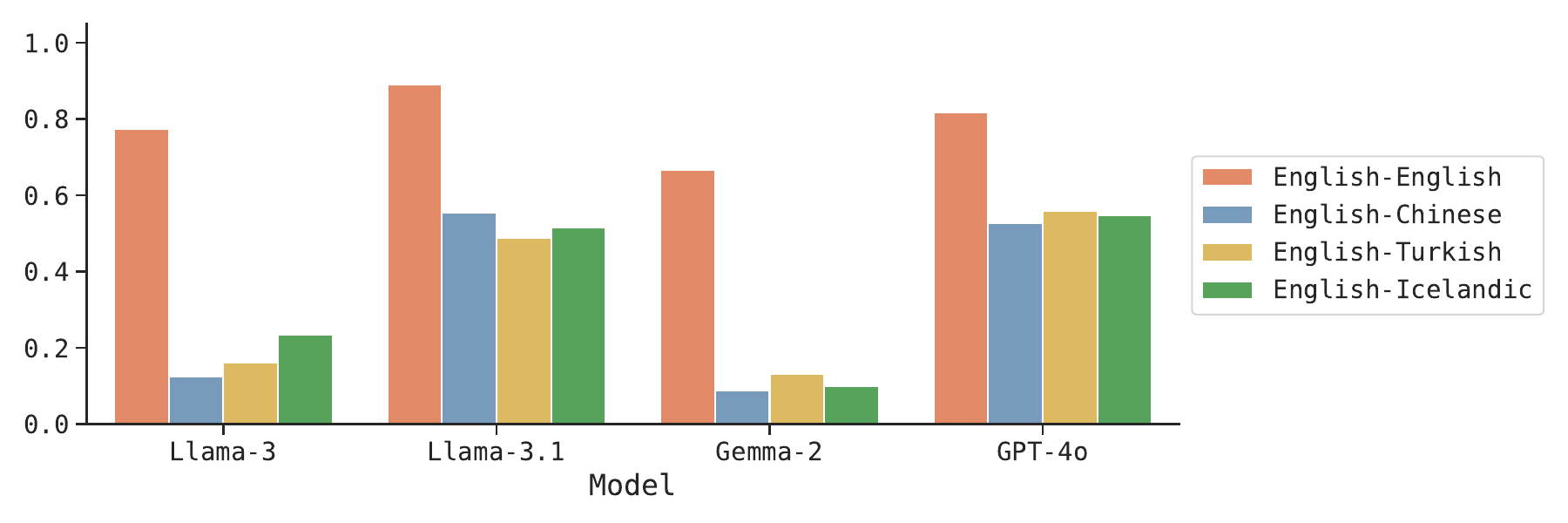} \\
    \caption{\label{fig:more-languages} Conflict detection ratio over cross-modal contexts with Icelandic and Turkish.}
\end{figure}

Figure~\ref{fig:more-languages} shows the results of conflict detection over cross-modal contexts containing Icelandic and Turkish.

\end{document}